%% file: acl2021.tex
\documentclass[11pt,a4paper]{article}
\usepackage[hyperref]{acl2021}
\usepackage{times}
\usepackage{latexsym}

\usepackage{microtype}

\usepackage{bm}
\usepackage{siunitx}
\usepackage{hyperref}
\usepackage{booktabs}
\usepackage{color}
\usepackage{graphicx}
\usepackage{amsmath}
\usepackage{amsthm}
\usepackage{amsfonts}
\usepackage{amssymb}
\usepackage{stmaryrd}
\usepackage{placeins}
\usepackage{multirow}
\DeclareMathOperator*{\bigtimes}{\vartimes}

\aclfinalcopy 
\def\aclpaperid{3760} 

\title{Exploring Discourse Structures for Argument Impact Classification}

\author{Xin Liu$^1$\thanks{\quad This work was done when Xin Liu was an intern at Huawei Noah’s Ark Lab.
} 
  \qquad \bf{Jiefu Ou}$^1$ 
  \qquad \bf{Yangqiu Song}$^1$
  \qquad \bf{Xin Jiang}$^2$ \\
$^1$Department of CSE, the Hong Kong University of Science and Technology\\
$^2$Huawei Noah’s Ark Lab \\
  \texttt{xliucr@cse.ust.hk}
  \ \texttt{jouaa@connect.ust.hk} \\
  \ \texttt{yqsong@cse.ust.hk}
  \ \texttt{jiang.xin@huawei.com}
}

\date{}

\begin{document}
\maketitle
\begin{abstract}
    Discourse relations among arguments reveal logical structures of a debate conversation.
    However, no prior work has explicitly studied how the sequence of discourse relations influence a claim's impact.
    This paper empirically shows that the discourse relations between two arguments along the context path are essential factors for identifying the persuasive power of an argument.
    We further propose \textsc{DisCOC} to inject and fuse the sentence-level structural discourse information with contextualized features derived from large-scale language models.
    Experimental results and extensive analysis show that the attention and gate mechanisms that explicitly model contexts and texts can indeed help the argument impact classification task defined by \citet{durmus2019the}, and discourse structures among the context path of the claim to be classified can further boost the performance.
\end{abstract}

\input{sections/introduction}
\input{sections/background}

\input{sections/discourse}
\input{sections/experiments}
\input{sections/related_work}
\input{sections/conclusion}

\section*{Acknowledgements}
This paper was supported by the NSFC Grant (No. U20B2053) from China, the Early Career Scheme (ECS, No. 26206717), the General Research Fund (GRF, No. 16211520), and the Research Impact Fund (RIF, No. R6020-19 and No. R6021-20) from the Research Grants Council (RGC) of Hong Kong, with special thanks to the Huawei Noah's Ark Lab for their gift fund.

\clearpage

\bibliographystyle{acl_natbib}
\bibliography{acl2021}

\clearpage

\input{sections/appendix}

\end{document}

%% file: sections/introduction.tex
\section{Introduction}

It is an interesting natural language understanding problem to identify the impact and the persuasiveness of an argument in a conversation.
Previous works have shown that many factors can affect the persuasiveness prediction, ranging from textual and argumentation features~\citep{wei2016post}, style factors~\citep{Baff2020Analyzing},
to the traits of source or audience~\citep{durmus2018exploring,Durmus2019Modeling,Shmueli2019Detecting}.
Discourse relations, such as \textit{Restatement} and \textit{Instantiation}, among arguments reveal logical structures of a debate conversation.
It is natural to consider using the discourse structure to study the argument impact.

\begin{figure}[!t]
    \centering
    \includegraphics[width=\linewidth]{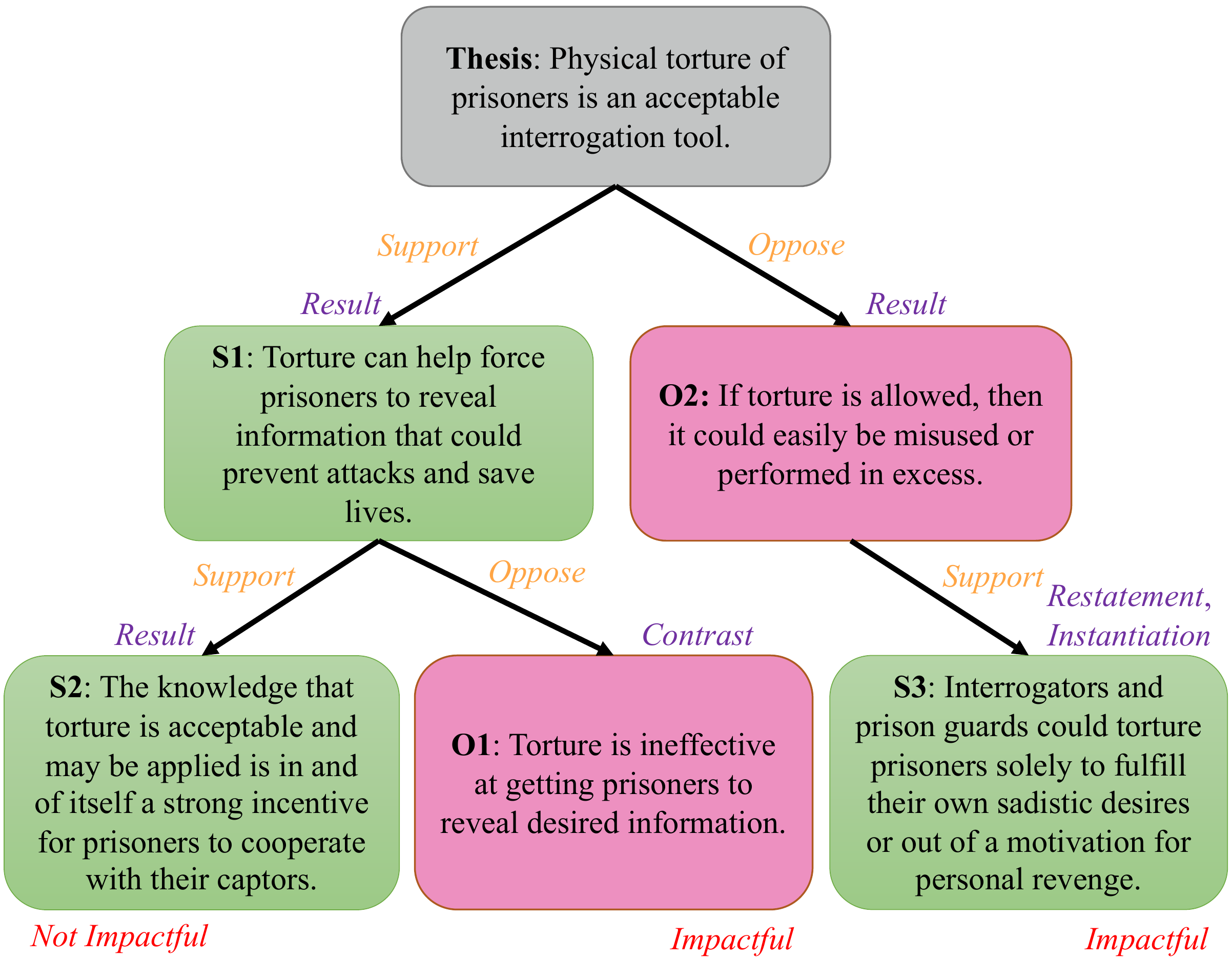}
    \vspace{-0.3cm}
    \caption{Example of an argument tree from \textit{Kialo}. Stances, impact labels, and discourse relations are annotated in orange, red, and violet respectively.}\label{fig:kialo}
    \vspace{-0.4cm}
\end{figure}

\citet{durmus2019the} initiated a new study of the influence of discourse contexts on determining argument quality by constructing a new dataset \textit{Kialo}.
As shown in Figure~\ref{fig:kialo}, it consists of arguments, impact labels, stances where every argument is located in an argument tree for a controversial topic.
They argue contexts reflect the discourse of arguments and conduct experiments to utilize historical arguments.
They find BERT with flat context concatenation is the best, but discourse structures are not easily captured by this method because it is difficult to reflect implicit discourse relations by the surface form of two arguments~\cite{Prasd2008The,LinKN09,XueNPPBR15,LanWWNW17,VariaHC19}.
Therefore, there is still a gap to study how discourse relations and their sequential structures or patterns affect the argument impact and persuasiveness prediction.


In this paper, we acquire discourse relations for argument pairs with the state-of-the-art classifier for implicit discourse relations.
Then we train a BiLSTM whose input is the sequence of discourse relations between two adjacent arguments to predict the last argument's impact, and the performance is comparable to that of a BiLSTM on raw text.
This indicates that a sequence of discourse relations is one of the essential factors for identifying the persuasive power of an argument.
Based on this intuition, 
we further propose a new model called \textsc{DisCOC} (\textbf{Dis}course \textbf{C}ontext \textbf{O}riented \textbf{C}lassifier) to explicitly produce discourse-dependent contextualized representations, fuse context representations in long distances, and make predictions.
By simple finetuning, our model beats the backbone RoBERTa~\cite{liu2019roberta} over 1.67\% and previous best model BERT over 2.38\%.
Extensive experiments show that \textsc{DisCOC} results in steady increases when longer context paths with discourse structures, e.g., stances and discourse relations, are provided.
On the contrary, encoders with full-range attentions are hard to capture such interactions, and narrow-range attentions cannot handle complex contexts and even become poisoned.

Our contributions can be highlighted as follows:

1.  To the best of our knowledge, we are the first to explicitly analyze the effect of discourse among contexts and an argument on the persuasiveness.

2. We propose a new model called \textsc{DisCOC} to utilize attentions to imitate recurrent networks for sentence-level contextual representation learning.

3. Fair and massive experiments demonstrate the significant improvement; detailed ablation studies prove the necessities of modules.

4.  Last, we discover distinct discourse relation path patterns in a machine learning way and conduct consistent case studies.

Code is publicly released at \url{https://github.com/HKUST-KnowComp/DisCOC}.

%% file: sections/background.tex
\section{Argument Tree Structure}

\subsection{Overview}
\textit{Kialo} dataset is collected by \citet{durmus2019the}, which consists of 47,219 argument claim texts from \textit{kialo.com} for 741 controversial topics and corresponding impact votes.
Arguments are organized as tree structures, where a tree is rooted in an argument thesis, and each node corresponds to an argument claim.
Along a path of an argument tree, every claim except the thesis was made to either support or oppose its parent claim and propose a viewpoint. 
As shown in Figure~\ref{fig:kialo}, an argument tree is rooted at the \textbf{thesis} ``Physical torture of prisoners is an acceptable interrogation tool.''.
There is one claim to support this thesis (\textbf{S1} in green) and one to oppose it (\textbf{O2} in fuchsia).
Moreover, \textbf{S1} is supported by its child claim \textbf{S2} and opposed by \textbf{O1}, and \textbf{S3} holds the same viewpoint of \textbf{O2}.

\subsection{Claim and Context Path}
As each claim was put in view of all its ancestral claims and surrounding siblings, the audience evaluated the claim based on how timely and appropriate it is.
Therefore, the context information is of most interest to be discussed and researched in the \textit{Kialo} dataset. 
We define that 
a \textbf{claim} denoted as $C$ is the argumentative and persuasive text to express an idea for the audience, 
and a \textbf{context path} of a claim of length $l$ is the path from the ancestor claim to its parent claim, denoted as $(C^0, C^1, \cdots, C^{l-1})$ where $C^{l-1}$ is the parent of $C$.
For simplicity, we may use $C^l$ instead of $C$ without causing ambiguity.
The longest path of $C$ starts from the thesis.
Statistically, the average length of the longest paths is 3.5.



\begin{table}[!t]
    \small
    \centering
    \begin{tabular}{c|c|c|c}
        \toprule
        Stance / Impact & Train & Validation & Test \\
        \midrule
        \textit{Pro} & 9,158 & 1,949 & 1,953 \\
        \textit{Con} & 8,695 & 1,873 & 1,891 \\
        \hline
        \textit{Impactful} & 3,021 & 641 &  646 \\
        \textit{Medium Impact} & 1,023  & 215 & 207 \\
        \textit{Not Impactful} & 1,126 & 252 & 255 \\
        \bottomrule
    \end{tabular}
    \vspace{-0.2cm}
    \caption{Statistics of stances and impact labels in the training, validation, and test data.}
    \vspace{-0.2cm}
    \label{table:stat_label}
\end{table}

\subsection{Argument Stance}
In a controversial topic, each argument claim except the thesis would have a stance, whether to support or oppose the argument thesis or its parent claim.
In \textit{Kialo}, users need to directly add a \textbf{stance} tag (\textit{Pro} or \textit{Con}) to show their agreement or disagreement about the chosen parent argument when they post their arguments.
We use $s^i$ to denote the stance whether $C^i$ is to support or oppose its parent $C^{i-1}$ when $i \geq 1$.
The statistics of these stances are shown in Table~\ref{table:stat_label}.

\begin{table*}[!ht]
    \small
    \centering
    \begin{tabular}{c|c|c|c|c|c|c|c}
        \toprule
        Discourse Relations & \textit{Reason} & \textit{Conjunction} & \textit{Contrast} & \textit{Restatement} & \textit{Result} & \textit{Instantiation} & \textit{Chosen Alternative} \\
        \hline
        Numbers & 6,559 & 6,421 & 5,718 & 5,343 & 1,355 & 99 & 23 \\
        \bottomrule
    \end{tabular}
    \vspace{-0.2cm}
    \caption{Statistics of predicted discourse relations.}
    \vspace{-0.2cm}
    \label{table:stat_discourse}
\end{table*}

\subsection{Impact Label}
After reading claims as well as the contexts, users may agree or disagree about these claims.
The \textbf{impact vote} for each argument claim is provided by users who can choose from 1 to 5.
\citet{durmus2019the} categorize votes into three impact classes (\textit{Not Impactful}, \textit{Medium Impact}, and \textit{Impactful}) based on the agreement and the valid vote numbers to reduce noise.
We can see the overall distribution from Table~\ref{table:stat_label}.
The argument impact classification is defined to predict the impact label $y$ of $C$ given the claim text $C$ and its corresponding context path $(C^0, C^1, \cdots, C^{l-1})$.

%% file: sections/discourse.tex

\section{Discourse Structure Analysis}

\subsection{Argument Impact from the Perspective of Discourse}
\label{sec:disco_analysis}
As paths under a controversial topic are strongly related to \textit{Comparison} (e.g., \textit{Contrast}), \textit{Contingency} (e.g., \textit{Reason}), \textit{Expansion} (e.g., \textit{Restatement}), and \textit{Temporal} (e.g., \textit{Succession}) discourse relations~\cite{Prasd2008The}, 
we model the discourse structures from a view of discourse relations. The first step is to acquire discourse relation annotations.
BMGF-RoBERTa~\cite{liu2020bmgfroberta} is the state-of-the-art model proposed to detect implicit discourse relations from raw text.
In the following experiments, we use that as our annotation model to predict discourse relation distributions for each adjacent claim pair.

Specifically, for a given argument claim $C^l$ and its context path $(C^0, C^1, \cdots, C^{l-1})$, we denote $p_{\text{disco}}(C^l) = (r^1, r^2, \cdots, r^l)$ as a \textbf{discourse relation path} such that $r^i \in \mathcal{R}$ indicates the discourse relation between $C^{i-1}$ and $C^i$ when $i \geq 1$.
In this work, we adopt the 14 discourse relation senses in CoNLL2015 Shared Task~\cite{XueNPPBR15} as $\mathcal{R}$. 
And we also define the corresponding \textbf{distributed discourse relation path} to be $p_{\text{dist}}(C^l) = (\bm{d}^1, \bm{d}^2, \cdots, \bm{d}^l)$ such that $\bm{d}^i = F(C^{i-1}, C^{i})$ is the predicted discourse relation distribution between claims $C^{i-1}$ and $C^{i}$ ($i \geq 1$) by a predictive model $\mathcal{F}$.
In experiments, $\mathcal{F}$ is BMGF-RoBERTa\footnote{The official open-source code is at \url{https://github.com/HKUST-KnowComp/BMGF-RoBERTa}.
We train such a classifier on CoNLL2015 Shared Task training data, and achieve 57.57\% accuracy on the test set.}.
8 out of 14 relations appear in the predictions, and the statistics of 7 frequent predictions are shown in Table~\ref{table:stat_discourse}.

\begin{figure}[!t]
    \centering
    \includegraphics[width=0.99\linewidth]{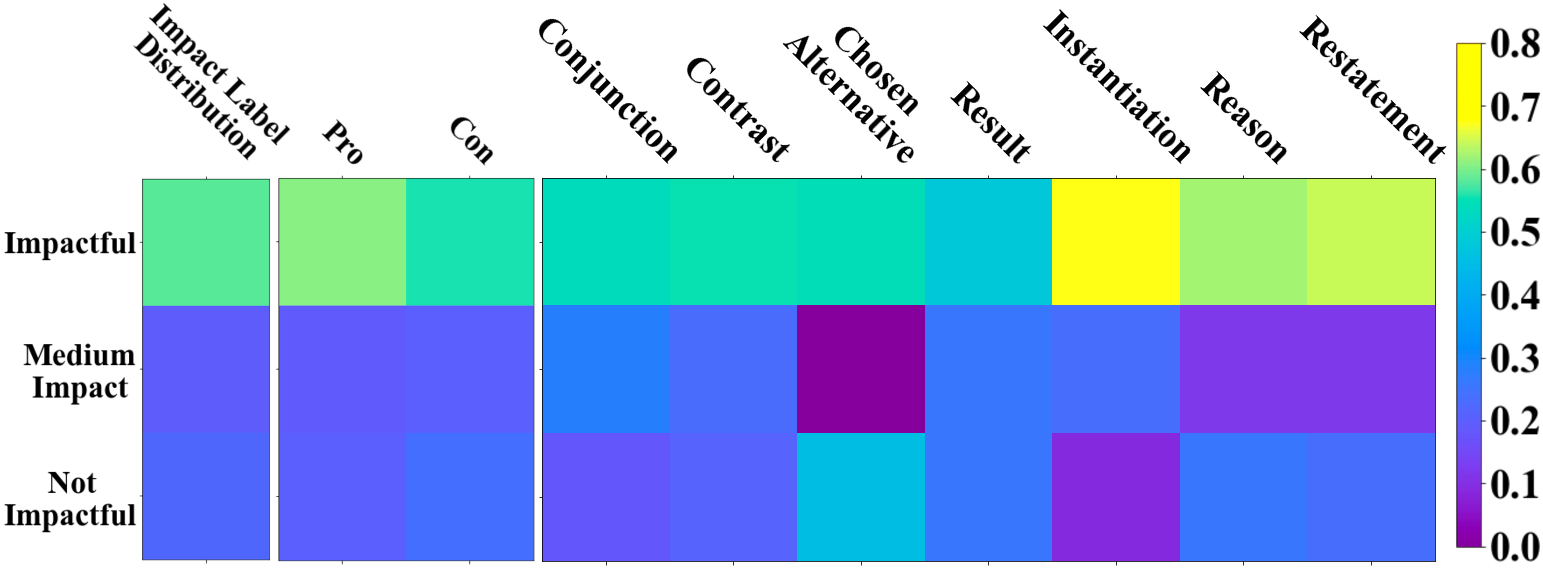}
    \vspace{-0.3cm}
    \caption{Impact label distributions, the correlations between labels and stances, and the correlations between labels and discourse relations. Normalization is applied to the columns.}\label{fig:corr}
    \vspace{-0.2cm}
\end{figure}

As discourse contexts would affect the persuasive power of claims, we first discover the correlations between impacts and stances as well as correlations between impacts and discourse relations, illustrated in Figure~\ref{fig:corr}.
From the label distribution and correlations, we find there are some clear trends: 
1) Stances have little influence on argument impact, but discourse relations do.
Correlations indicate that it is the contents instead of standpoints that contribute to potential impacts;
2) It is a smart choice to show some examples to convince others because \textit{Instantiation} is more relevant to \textit{Impactful} than any other relations;
3) Similarly, explaining is also helpful to make voices outstanding;
4) \textit{Restatement} is also positively correlated with \textit{Impactful} so that we can also share our opinions by paraphrasing others' viewpoints to command more attention. On the contrary, \textit{Chosen Alternative} is a risky method because the audience may object.



\begin{figure}[t]
    \centering
    \includegraphics[width=0.85\linewidth]{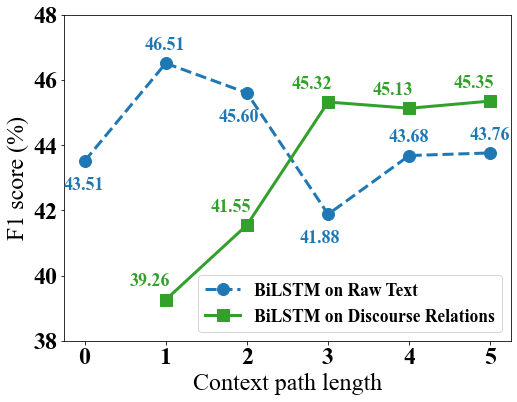}
    \vspace{-0.2cm}
    \caption{Performance of BiLSTM on discourse relations and BiLSTM on raw text.}\label{fig:lstm}
    \vspace{-0.3cm}
\end{figure}

\label{discourse_modeling}
To investigate the role of discourse relations in impact analysis, we design a simple experiment that a single-layer BiLSTM followed by a 2-layer MLP with batch normalization predicts the impact by utilizing the distributed discourse relation path $p_{\text{dist}}(C^l)$.
For the purposes of comparison and analysis, we build another BiLSTM on the raw text. Each claim has \text{[BOS]} and \text{[EOS]} tokens to clarify boundaries and we use 300-dim 
pretrained GloVe word embeddings~\cite{Pennington2014Glove} and remain them fixed.
We set different thresholds for context path lengths so that we can control how many discourse relations or contexts are provided.
From Figure~\ref{fig:lstm}, discourse features can result in comparable performance, especially when longer discourse paths are provided. Instead, the model with raw text gets stuck in complex contexts.

\begin{figure*}[t]
    \raggedleft
    \includegraphics[width=0.95\linewidth]{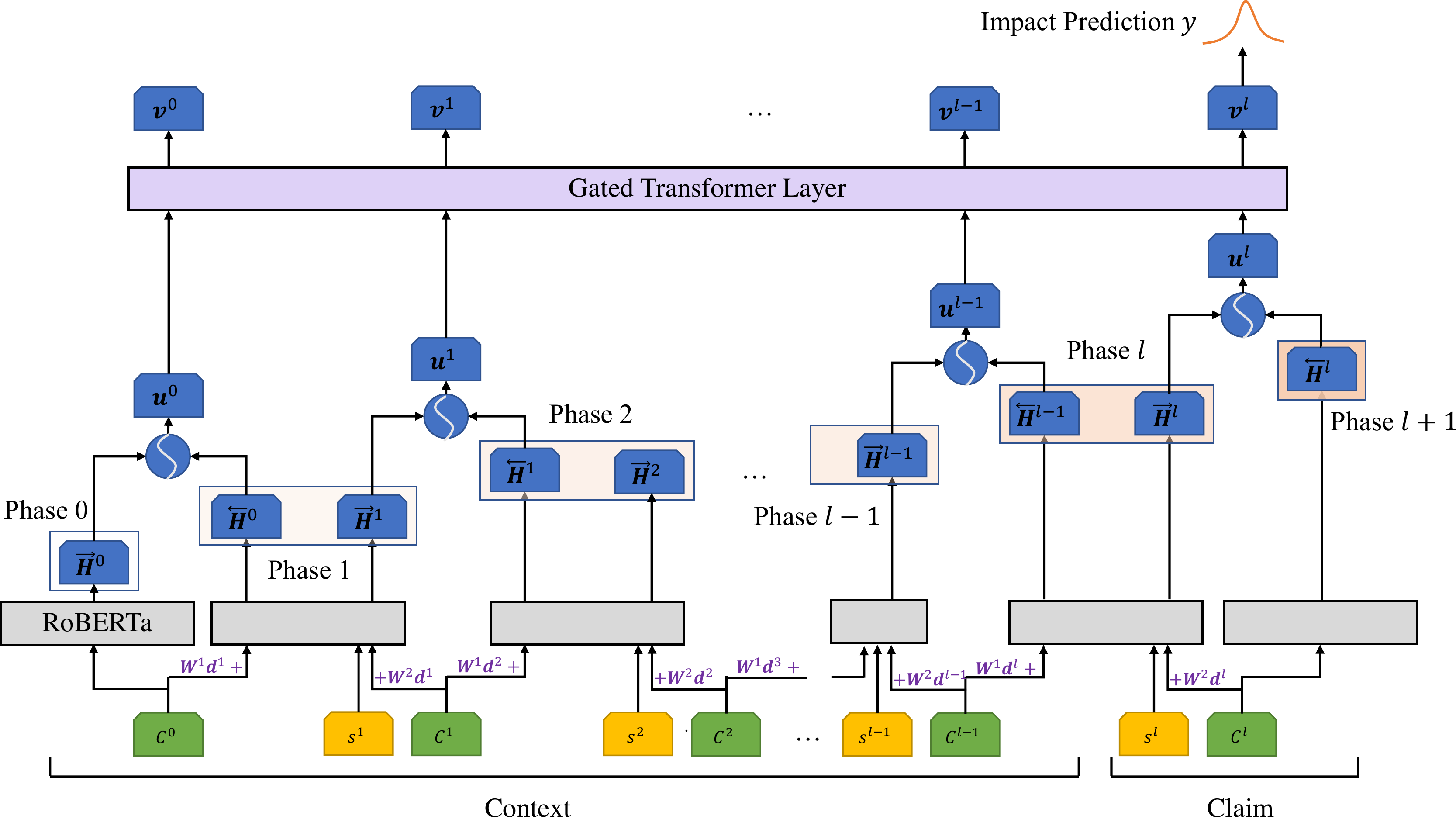}
    \vspace{-0.2cm}
    \caption{The architecture of \textsc{DisCOC}.
    $s^i$ refers to the stance between $C^{i-1}$ and $C^{i}$, $\bm{d}^{i}$ is the discourse relation distribution obtained from $\mathcal{F}(C^{i-1}, C^{i})$. Gray boxes represent the RoBERTa encoder and the violet is a gated transformer layer.
    \text{[CTX]}, \text{[CLS]}, and \text{[SEP]} are omitted in this figure.}\label{fig:arch}
    \vspace{-0.2cm}
\end{figure*}

\subsection{Discourse Context Oriented Classifier}
It is generally agreed that the informative context can help understand the text to be classified. 
However, it is still unclear how to determine whether a context is helpful. 
One drawback of a broader context is the increasing ambiguity, especially in the scenario of the argument context path from different users like the results shown in Figure~\ref{fig:lstm}.
Take claims in Figure~\ref{fig:kialo} for example,
\textbf{S1} and \textbf{O2} give two different consequences to support or oppose the \textbf{thesis}. And \textbf{O1} objects \textbf{S1} by a contrast conclusion.
It is hard to build a connection between the \textbf{thesis} and \textbf{O1} if \textbf{S1} is not given because it is challenging to build a connection between ``reveal desired information'' with ``interrogation tool'' without a precondition ``Torture can help force prisoners to reveal information''.
On the contrary, \textbf{thesis} and \textbf{S2} are still compatible as \textbf{S2} is also a kind of result.
Hence, a recurrent model with the gating mechanism that depicts pair-wise relations and passes to the following texts makes more sense.

LSTM has gates to decide whether to remember or forget during encoding, but it cannot handle long-range information with limited memory.
Recently, transformer-based encoders have shown remarkable performance in various complicated tasks.
These models regard sequences as fully connected graphs to learn the correlations and representations for each token.
People assume that transformers can learn whether two tokens are relevant and how strong the correlation is by back-propagation.
Table~\ref{table:attention} illustrates different possible ways to aggregation context information.
Transformer~\cite{VaswaniSPUJGKP17} and BERT~\cite{devlin2019bert} adopt full-range attentions while TransformerXL~\cite{DaiYYCLS19} and XLNet~\cite{YangDYCSL19} regard historical encoded representations as memories to reuse hidden states.
SparseTransformer~\cite{ChildGRS2019}, in the opposite direction, stacks hundreds of layers by narrow the attention scope by sparse factorization.
Information can still spread after propagations in several layers.
Inspired by these observations, we design \textsc{DisCOC} (\textbf{Dis}course \textbf{C}ontext \textbf{O}riented \textbf{C}lassifier) to capture contextualized features by localized attentions and imitate recurrent models to reduce noises from long distance context.
As shown in Figure~\ref{fig:arch}, \textsc{DisCOC} predicts the argument impact through three steps.

\begin{table}[!t]
    \small
    \centering
    \setlength{\tabcolsep}{3pt}
    \begin{tabular}{c|c|c|c}
        \toprule
        Attention & Representative & Query & Key \& Value \\
        \hline
        Full & BERT & $C^i$ & $C^0, \cdots, C^l$ \\
        Memory & XLNet & $C^i$ & ($C^0, \cdots, C^{i-1}$)  \\
        Context & SparseTransformer & $C^i$ & $C^{i-1}$ \\
        \bottomrule
    \end{tabular}
    \vspace{-0.2cm}
    \caption{Different attention mechanisms. The Memory attention freezes the historical representations so that gradients of $C^{i}$ would not propagate to the memory ($C^0, \cdots, C^{i-1}$).}
    \vspace{-0.3cm}
    \label{table:attention}
\end{table}

\subsubsection{Adjacent Claim Pair Encoding}
A difficult problem in such an argument claim tree is the noise in irrelevant contexts. A claim is connected to its parent claim because of a supporting or opposing stance, but claims in long distances are not high-correlated.
Based on this observation, \textsc{DisCOC} conduct word-level representations by encoding claim pairs instead of the whole contexts.

Given a claim $C^l$ and its context path $(C^0, C^1, \cdots, C^{l-1})$, all adjacent pairs are coupled together, i.e., $(C^0, C^1), \cdots, (C^{l-1}, C^{l})$.
We can observe that each claim appears twice except the first and the last.
Next, each pair $(C^{i-1}, C^{i})$ is fed into the RoBERTa encoder to get the contextualized word representations. $C^0$ and $C^l$ are also encoded separately so that each claim has been encoded twice.
We use $\overrightarrow{\bm{H}}^{i}$ to denote the encoded word representations of $C^{i}$ when this claim is encoded with its parent $C^{i-1}$, or when it is computed alone as $C^0$.
Similarly, $\overleftarrow{\bm{H}}^{i}$ is the representations when encoding $(C^{i}, C^{i+1})$, or when it is fed as $C^{l}$.

The encoding runs in parallel but we still use the term \textit{phase} to demonstrate for better understanding.
In 0-th phase, RoBERTa outputs $\overrightarrow{\bm{H}}^{0}$.
One particular relationship between a parent-child pair is the stance, and we insert the one special token \text{[Pro]} or \text{[Con]} between them.
It makes the sentiment and viewpoint of the child claim more accurate.
On the other hand, discourse relations can also influence impact prediction, as reported in Section~\ref{sec:disco_analysis}.
However, discourse relations are not mutually exclusive, let alone predictions from BMGF-RoBERTa are not precise.
Thus, we use the relation distributions as weights to get sense-related embeddings over 14 relations.
We add additional $\bm{W}^1 \bm{d}^i$ for the parent and $\bm{W}^2 \bm{d}^i$ for the child except position embeddings and segment embeddings, where $\bm{d}^{i}$ is predicted discourse relation distribution for $(C^{i-1}, C^{i})$, $\bm{W}^1$ and $\bm{W}^2$ are trainable transformations for parents and children.
Hence, RoBERTa outputs $\overleftarrow{\bm{H}}^{i-1}$ and $\overrightarrow{\bm{H}}^{i}$ with the concatenation of two claims, \text{[CTX]} $C^{i-1}$ \text{[SEP]} \text{[CLS]} $s^i$ $C^{i}$ \text{[SEP]} in the $i$-th phase ($i \in \{1, 2, \cdots, l\}$), where \text{[CTX]} is a special token to indicate the parent claim and distinguish from \text{[CLS]}. Its embedding is initialized as a copy embedding of \text{[CLS]} but able to update by itself.
And $\overleftarrow{\bm{H}}^{l}$ is computed by self-attention with no context in the last phase.
In the end, each claim $C^{i}$ has two contextualized representations $\overleftarrow{\bm{H}}^{i}$ and $\overrightarrow{\bm{H}}^{i}$ with limited surrounding context information.

\subsubsection{Bidirectional Representation Fusion}
As claim representations $\{\overleftarrow{\bm{H}}^{i}\}$ and $\{\overrightarrow{\bm{H}}^{i}\}$ from RoBERTa are not bidirectional, we need to combine them and control which of them matters more.
The gated fusion~\cite{liu2020bmgfroberta} has been shown of a better mixture than the combination of multi-head attention and layer normalization.
We use it to maintain the powerful representative features and carry useful historical context information:
\begin{align}
     \hat{\bm{H}^{i}} &= \text{MultiHead}(\overleftarrow{\bm{H}}^{i}, \overrightarrow{\bm{H}}^{i}, \overrightarrow{\bm{H}}^{i}) \\
     \bm{A}_j &= \text{Sigmoid}(\bm{W}^a [\overleftarrow{\bm{H}}^{i}, \hat{\bm{H}^{i}}]_j + \bm{b}^a) \\
     \bm{U}^{i} &= \bm{A} \odot \hat{\bm{H}^{i}} + (1 - \bm{A}) \odot \overleftarrow{\bm{H}}^{i},
\end{align}
where \text{MultHead} is the multi-head attention operation~\cite{VaswaniSPUJGKP17} whose query is $\overleftarrow{\bm{H}}^{i}$ and key \& value is $\overrightarrow{\bm{H}}^{i}$, $\bm{A}_j$ is the fusion gate for the $j$-th word embedding, $[\cdots]$ is the concatenation, $\odot$ is the element product operation, and $\bm{W}^a$ and $\bm{b}^a$ are trainable matrix and bias for fusion gating.
There are two reasons why using $\overleftarrow{\bm{H}}^{i}$ as the key of the multi-head attention: 1) \text{[CLS]} exists in the $\overleftarrow{\bm{H}}^{i}$ while the replaced token \text{[CTX]} appears in $\overrightarrow{\bm{H}}^{i}$ when $i \neq 0$; 2) The position ids start from 0 when computing $\overleftarrow{\bm{H}}^{i}$.
The fused \text{[CLS]} token embedding $\bm{u}^{i}$ is selected to represent the whole claim.

\subsubsection{Context Path Information Gathering}

After extracting sentence-level claim representations $\bm{u}^0, \bm{u}^1, \cdots, \bm{u}^l$, a transformer layer is used to gather longer-range context representations.
The transformer layer includes a position embedding layer to provide sinusoid positional embeddings, a gated multi-head attention layer, a feed-forward network, and a layer normalization.
The position embedding layer in \textsc{DisCOC} is different from that in the vanilla Transformer because it generates position ids in a reversed order, i.e. $l, l-1, \cdots, 0$.
The reversed order is helpful to model the contexts of variable length because the claim to be classified has the same position embedding.
We also choose a gate to maintain the scale instead of using a residual connection.
The gated transformer can generate meaningful representations because each claim can attend any other claims and itself.
On the other hand, it perfectly fits the pair-wise encoding that imitates the recurrent networks to reduce the noise in irrelevant contexts and enhance the nearest context's correlations.
For example, in Figure~\ref{fig:kialo}, \textbf{S2} is predicted as a result of \textbf{S1} (with a probability of 39.17\%) and a restatement (with a probability of 19.81\%), and \textbf{S1} is also a result of \textbf{thesis} (with a probability of 70.57\%).
Consequently, \textbf{S2} is high-relevant to the \textbf{thesis} as a potential result if ``physical torture is acceptable'', which can be captured by \textsc{DisCOC}.
Finally, a 2-layer MLP with batch normalization is applied to $\bm{v}^l$ of the last claim to predict its impact.

%% file: sections/experiments.tex
\section{Experiments}

\subsection{Baseline Models}

\paragraph{Majority.} The baseline simply returns \textit{Impactful}.
\paragraph{SVM.} \citet{durmus2019the} created linguistic features for a SVM classifier, such as named entity types, POS tags, special marks, tf-idf scores for n-grams, etc. We report the result from their paper.
\paragraph{HAN.} HAN~\cite{YangYDHSH16} computes document vectors in a hierarchical way of encoding and aggregation. 
We replace its BiGRU with BiLSTM for the sake of comparison. And we also extend it with pretrained encoders and transformer layers.
\paragraph{Flat-MLMs.} Pretrained masked languages, e.g., RoBERTa, learn word representations and predict masked words by self-attention. We use these encoders to encode the flat context concatenation like \text{[CTX]} $C^{0}$ \text{[SEP]} \text{[CTX]} $\cdots$ \text{[CTX]} $C^{l-1}$ \text{[SEP]} as Segment A and  \text{[CLS]} $C^l$ \text{[SEP]} as Segment B. After getting \text{[CTX]} and \text{[CLS]} representations, a gated transformer layer and a MLP predict impacts.
As for XLNet, we follow its default setting so that \text{[CTX]} and \text{[CLS]} are located at the end of claims.
\paragraph{Interval-MLMs.} Flat-MLMs regard the context path as a whole segment and ignore the real discourse structures except the adjacency, e.g., distances between two claims are missing.
We borrow the idea from BERT-SUM~\cite{LiuL19}:
segment embeddings of $C^{i}$ are assigned depending on whether the distance to $C^l$ is odd or even.
\paragraph{Context-MLMs.} We also compare pretrained encoders with context masks. A context mask is to localize the attention scope from the previous to the next. That is, $C^i$ can attends words in $C^{i-1}$ and $C^{i+1}$ except for itself if $1 \leq i < l$; $C^0$ can only attend $C^0, C^1$, and $C^l$ can only attend $C^{l-1}, C^{l}$.
\paragraph{Memory-MLMs.} XLNet utilizes memory to extend the capability of self-attention to learn super long historical text information.
We also extend Flat-MLMs under this setting.

\begin{table*}[!h]
    \centering
    \small
    \vspace{-0.05cm}
    \begin{tabular}{l|l|l|l}
    \toprule
    \multicolumn{1}{c|}{Model} &  \multicolumn{1}{c|}{Precision} &  \multicolumn{1}{c|}{Recall} &  \multicolumn{1}{c}{F1} \\ 
    \midrule
    Majority & 19.43 & 33.33 & 24.55 \\
    SVM~\cite{durmus2019the} & \bf{65.67} & 38.58 & 35.42 \\
    BiLSTM & 46.94 $\pm$ 1.08** & 46.64 $\pm$ 0.71** & 46.51 $\pm$ 1.11** \\
    \hline
    HAN-BiLSTM & 51.93 $\pm$ 1.37** & 49.08 $\pm$ 1.52** & 50.00 $\pm$ 1.49** \\
    HAN-BERT & 53.72 $\pm$ 0.80** & 53.45 $\pm$ 0.51** & 53.46 $\pm$ 0.47** \\
    HAN-RoBERTa & 55.71 $\pm$ 1.12** & 55.95 $\pm$ 0.90** & 55.49 $\pm$ 0.62** \\
    HAN-XLNet & 53.91 $\pm$ 0.96** & 55.56 $\pm$ 1.59** & 54.53 $\pm$ 1.22** \\
    \hline
    BERT~\cite{durmus2019the} & 57.19 $\pm$ 0.92 & 55.77 $\pm$ 1.05** & 55.98 $\pm$ 0.70** \\
    Flat-BERT & 57.34 $\pm$ 1.56 & 57.07 $\pm$ 0.74* & 56.75 $\pm$ 0.82** \\
    Flat-RoBERTa & 58.11 $\pm$ 1.34 & 56.40 $\pm$ 0.61** & 56.69 $\pm$ 0.63** \\
    Flat-XLNet & 55.86 $\pm$ 1.74* & 56.20 $\pm$ 1.17** & 55.57 $\pm$ 0.95** \\
    \hline
    Interval-BERT & 55.56 $\pm$ 2.03* & 55.52 $\pm$ 1.44** & 55.34 $\pm$ 1.50** \\
    Interval-RoBERTa & 58.31 $\pm$ 0.89 & 56.46 $\pm$ 1.44* &	56.61 $\pm$ 1.24* \\
    Interval-XLNet & 57.54 $\pm$ 0.50 & 56.78 $\pm$ 1.63* & 56.52 $\pm$ 1.00** \\
    \hline
    Context-BERT & 54.96 $\pm$ 0.93** & 56.09 $\pm$ 0.83** & 55.44 $\pm$ 0.83** \\
    Context-RoBERTa & 57.28 $\pm$ 0.97 & 55.29 $\pm$ 0.26** & 55.83 $\pm$ 0.54** \\
    Context-XLNet & 54.56 $\pm$ 0.71** & 56.28 $\pm$ 1.22** & 55.10 $\pm$ 0.72** \\
    \hline
    Memory-BERT & 54.33 $\pm$ 0.83** & 57.57 $\pm$ 0.67* & 55.22 $\pm$ 0.61** \\
    Memory-RoBERTa & 55.08 $\pm$ 0.89** & 55.55 $\pm$ 1.59** & 54.76 $\pm$ 1.38** \\
    Memory-XLNet & 55.44 $\pm$ 1.15** & 55.45 $\pm$ 1.25** & 54.91 $\pm$ 0.96** \\
    \hline
    \textsc{DisCOC} & 57.90 $\pm$ 0.70 & \bf{59.41 $\pm$ 1.41} & \bf{58.36 $\pm$ 0.52} \\
    \bottomrule
    \end{tabular}
    \vspace{-0.15cm}
    \caption{The averages and standard deviations of different models on the argument impact classification. The marker * refers to $p$-value $<$ 0.05 and the marker ** refers to $p$-value $<$ 0.001 in t-test compared with \textsc{DisCOC}.}
    \label{table:arg_impact}
    \vspace{-0.2cm}
\end{table*}

\subsection{Model Configuration and Settings}

We use pretrained base models~\footnote{BERT-base-uncased, RoBERTa-base, and XLNet-base-cased are downloaded from \url{huggingface.co}} in \textsc{DisCOC} and baselines.
We follow the same finetuning setting:
classifiers are optimized by Adam~\cite{Kingma2014Adam} with a scheduler and a maximum learning rate {2e-5}.
The learning rate scheduler consists of a linear warmup for the 6\% steps and a linear decay for the remaining steps. 
As for BiLSTM and HAN, the maximum learning rate is {1e-3}.
The hidden state dimension of linear layers, the hidden units of LSTM layers, and projected dimensions for attention are 128.
The number of the multi-head attention is set as 8.
Dropout is applied after each layer and the probability is 0.1.
We pick the best context path length $l$ for each model by grid search from 0 to 5 on validation data with the batch size of 32 in 10 epochs.
Each model runs five times.

\subsection{Argument Impact Classification}
\label{sec:arg_impact}

Table~\ref{table:arg_impact} shows experimental results of different models.
It is not surprising that neural models can easily beat traditional feature engineering methods in overall performance. But linguistic features still bring the highest precision.
We also observe a significant 3.49\% improvement with context vectors aggregating in HAN-BiLSTM compared with the simple BiLSTM.
This indicates that it is necessary to model contexts with higher-level sentence features.
Models with pretrained encoders benefit from representative embeddings, and HAN-RoBERTa achieves a gain of 5.49\%.
Flat context paths contain useful information to help detect the argument impact, but they also involve some noise from unrelated standpoints.
Interval segment embeddings do not reduce noise but make BERT confused.
It is counterintuitive that the segment embeddings depend on whether the distance is odd or even because BERT uses these for next sentence prediction.
Since XLNet uses relative segment encodings instead of segment embeddings, Interval-XNet is better than Flat-XLNet in all three metrics.
On the other hand, context masks bring another side effect for BERT, RoBERTa, and XLNet.
Although these masks limit the attention scope at first sight, distant word information is able to flow to words with the increment of transformer layers.
As a result, the uncertainty and attention bias increase after adding context masks.
The memory storing context representations is also not helpful.
The main reason is that the last claim's update signal can not be used to update previous context representations.
That is, Memory-models degenerate to models with frozen path features or even worth.
\textsc{DisCOC} that we proposed can capture useful contexts and fuse in a comprehensive manner.
Finally, \textsc{DisCOC} outperforms the second best model Flat-BERT over 1.61\% and its backbone Flat-RoBERTa over 1.67\%, the previous best model BERT by 2.38\%.

\subsection{Ablation Study}

\subsubsection*{Influence of the Context Path Length}

\begin{figure}[t]
    \centering
    \includegraphics[width=0.86\linewidth]{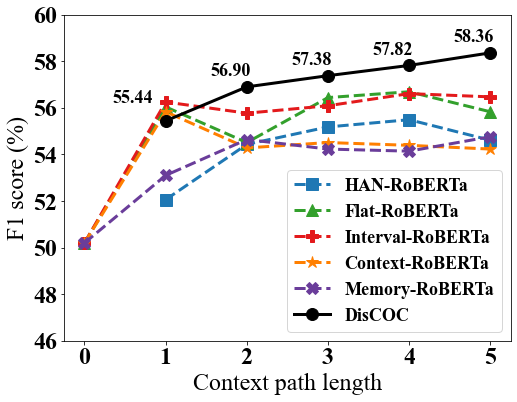}
    \vspace{-0.3cm}
    \caption{F1 scores of different models on varying the maximum path length.
    }\label{fig:path_len}
    \vspace{-0.2cm}
\end{figure}

Different claims have different contexts.
We only report the best performance with a fixed maximum context path length in Table~\ref{table:arg_impact}.
Figure~\ref{fig:path_len} shows F1 scores of models with different hyper-parameters. 
\textsc{DisCOC} always benefits from longer discourse contexts while other models get stuck in performance fluctuation.
Most models can handle one context claim, which is consistent with our idea of pair-wise encoding.
\textsc{DisCOC} has consistent performance gains; instead, other models cannot learn long-distance structures better.
Each token in Flat-RoBERTa and Interval-RoBERTa can attend all other tokens, and the two are the most competitive baselines.
However, Context-RoBERTa and Memory-RoBERTa limit the attention scope to the tokens of one previous claim, making models unable to make use of long-distance context information.


\begin{table}[!t]
    \centering
    \small
    \begin{tabular}{l|c|c|c}
    \toprule
    \multicolumn{1}{c|}{Model} &  \multicolumn{1}{c|}{Precision} & \multicolumn{1}{c|}{Recall} & \multicolumn{1}{c}{F1} \\ 
    \midrule
    \textsc{DisCOC} & 57.90  & 59.41 & \bf{58.36} \\
    \textsc{DisCOC} (E-BERT) & 57.84 & \bf{59.46} & 58.04 \\
    \textsc{DisCOC} (w/o StanceE) & \bf{58.68} & 58.12 & 57.74 \\
    \textsc{DisCOC} (w/o DiscoE) & 57.81 & 58.42 & 57.29 \\
    \textsc{DisCOC} (F-BiLSTM) & 58.58 & 57.87 & 57.72 \\
    \textsc{DisCOC} (F-Conv) & 58.20 & 58.53 & 57.82 \\
    \textsc{DisCOC} (w/o GTrans) & 56.04 & 54.71 & 54.78 \\
    \bottomrule
    \end{tabular}
    \vspace{-0.2cm}
    \caption{Ablation Studies of \textsc{DisCOC}.}
    \label{table:ablation}
    \vspace{-0.3cm}
\end{table}

\subsubsection*{RoBERTa vs. BERT}
As shown in Table~\ref{table:arg_impact}, there is little difference between the performance of RoBERTa variants and that of BERT variants.
We conduct the experiment for \textsc{DisCOC (E-BERT)} with BERT as the encoder reported in Table~\ref{table:ablation}.
Its performance has achieved a significant boost over 1.29\% despite the small gap between itself and \textsc{DisCOC}.

\begin{table*}[!t]
    \footnotesize
    \centering
    \begin{tabular}{c|c|c}
    \toprule
    Impactful & Medium Impact & Not Impactful \\ 
    \midrule
    Reason-Contrast & Conjunction-Reason & Restatement-Reason \\
    Restatement & Conjunction-Contrast & Contrast-Restatement \\
    Reason & Contrast-Conjunction & Chosen Alternative \\
    Restatement-Conjunction & Conjunction-Restatement & Restatement-Restatement \\
    Restatement-Contrast & Contrast-Contrast & Reason-Restatement \\
    Contrast-Instantiation & Contrast-Reason & Chosen Alternative-Reason \\
    Conjunction-Instantiation & Conjunction-Conjunction-Restatement & Contrast \\
    Restatement-Restatement & Conjunction-Restatement-Conjunction & Chosen Alternative-Conjunction \\
    Reason-Conjunction & Conjunction-Reason-Conjunction & Result-Reason \\
    Restatement-Result & Conjunction-Conjunction & Chosen Alternative-Restatement \\
    \bottomrule
    \end{tabular}
    \vspace{-0.2cm}
    \caption{Discourse path patterns that corresponding to the largest top 10 coefficients of the binary LR.}
    \vspace{-0.2cm}
    \label{table:disco_pattern}
\end{table*}

\subsubsection*{Are Stances and Discourse Senses Helpful?}

We also remove either the stance token embedding or the discourse sense embeddings from \textsc{DisCOC}.
The results in Table~\ref{table:ablation} suggest that both sides of structures are essential for modelling the correlation between the parent claim and the child claim.
By comparison, discourse sense embeddings are more vital.

\subsubsection*{Are Gated Transformers Necessary?}

We add a gated transformer layer to gather sentence-level vectors. Such gathering is necessary for the proposed framework
because each claim can only attend limited contexts.
BiLSTM and convolutions can also be used for this purpose, so we replace the gated transformer layer with a BiLSTM or a convolutional layer.
Moreover, we also remove it to make predictions by $\bm{u}^l$ directly.
The results in Table~\ref{table:ablation} show that the gated transformer is the irreplaceable part of \textsc{DisCOC} because it retains the contextualized representations and remains their scales.
Simple removing it hurts recall enormously.

\subsection{What Makes Claims Impactful?}
\subsubsection*{High-coefficient Discourse Relation Patterns}

We use Logistic Regression to mine several interesting discourse relation patterns. 
Detailed settings are described in Appendix~\ref{appendix:path_pattern}, and results including the most high-coefficient patterns are listed in Table~\ref{table:disco_pattern}.
We observe that some discourse relation path patterns are distinguishing for classifying individual impact labels.
\textit{Instantiation} is a typical relation that only occurs in the top patterns of \textit{Impactful}. 
Also, \textit{Restatement} is relatively frequent for 
\textit{Impactful} (5 of top 10), but it is the relation between the grandparent and the parent.
Providing additional resources (\textit{Restatement}-\textit{Result}) or objecting others' repetitions (\textit{Restatement}-\textit{Contrast}) can increase the persuasive power.
For the \textit{Medium Impact} class, its top 10 significant patterns are the longest on average.
That indicates some views are usually considered ordinary in complex structures.
\textit{Conjunction} is the dominant relation (8 of top 10) so that we are suggested to avoid to go along with others.
The case of \textit{Not Impactful} is a little clearer, 
in the sense that it has a unique relation \textit{Chosen Alternative} as one of the most significant patterns.
\textit{Restatement} also appears frequently, showing neither generalization, nor specification, nor paraphrasing of others' views can help make claims stand out.

\begin{table}[!t]
    \footnotesize
    \centering
    \begin{tabular*}{1.01\linewidth}{@{\extracolsep{\fill}}c|c|c}
    \toprule
    Discourse Patterns & \textsc{DisCOC} & \textsc{DisCOC} (w/o DiscoE) \\ 
    \midrule
    Reason-Contrast & \bf{65.56} & 43.33 \\
    Restatement & 56.63 & \bf{57.59} \\
    Reason & \bf{58.91} & 54.96 \\
    Conjunction-Reason & \bf{78.97} & 72.14 \\
    Conjunction-Contrast & \bf{80.64} & 66.17 \\
    Contrast-Conjunction & \bf{55.15} & 42.38 \\
    Restatement-Reason & \bf{38.00} & 37.35 \\
    Contrast-Restatement & 66.10 & \bf{76.24} \\
    Chosen Alternative & \bf{73.33} & 42.86 \\
    \hline
    All & \bf{59.04} & 58.06 \\
    \bottomrule
    \end{tabular*}
    \caption{F1 score differences between two best models on top 9 discourse relation patterns and all patterns.}
    \label{table:margins}
\end{table}

\subsubsection*{Case Study}
In Appendix~\ref{appendix:path_pattern}, we define $Pr(r^1, \cdots, r^l)$ as the joint probability to generate the discourse relation path $(r^1, \cdots, r^l)$ given the context $(C^0, C^1, \cdots, C^{l-1})$ and the claim $C^l$.
For example, the $Pr(\textit{Reason}, \textit{Contrast})$ is 56.59\% which corresponds to an \textit{Impactful} claim ``There is no evidence for this'' with its parent claim ``Our bodies know how to recognise and process current foods; changing them through genetic modification will create health issues''.
Furthermore, we find 5 of top 5 and 8 of top 10 are voted as \textit{Impactful} claims after sorting based on $Pr(\textit{Reason}, \textit{Contrast})$.
For a complex pattern \textit{Restatement}-\textit{Restatement} appearing in both top patterns of the \textit{Impactful} and the \textit{Not Impactful}, 3 cases with the maximum probabilities are \textit{Not Impactful} while the following 7 cases are \textit{Impactful}.
It is interesting that the thesis of the top 3 claims is the same discussion about an American politician.
There are 25 \textit{Impactful} claims and 22 \textit{Not Impactful} claims in this topic, 24 of which are restatements of their parent claims.
As for \textit{Restatement}-\textit{Reason}, the most top pattern of the \textit{Not Impactful}, we find 7 of the top 10 claims relevant to politics, 2 of them about globalization, and one food-related.
Therefore, there is no perfect answer in these quite controversial topics, and that is why \textit{Restatement} and \textit{Reason} appear frequently.

\subsubsection*{Empirical Results}
On the other hand, we check the performance of testing examples to verify the effectiveness of these discourse relation patterns.
We choose the best model of \textsc{DisCOC}, whose F1 score is 59.04\% as well as the best model of \textsc{DisCOC} (w/o DiscoE) whose F1 score is 58.06\%. We select testing examples with specific discourse patterns, and performance differences are shown in Table~\ref{table:margins}.
\textsc{DisCOC} benefits from 7 of the top 9 patterns and the performance margins are even more significant than the improvement of the overall results. Without giving discourse relation patterns, the model still has trouble capturing such implicit context influences.
Empirical results support our idea that implicit discourse relations could affect the persuasiveness.

%% file: sections/related_work.tex
\section{Related Work}

There is an increasing interest in computational argumentation to evaluate the qualitative impact of arguments based on corpus extracted from Web Argumentation sources such as CMV sub-forum of Reddit~\cite{Tan2016Winning}. 
Studies explored the importance and effectiveness of various factors on determining the persuasiveness and convincingness of arguments, such as surface texture, social interaction and argumentation related features~\cite{wei2016post}, characteristics of the source and audience~\cite{Durmus2019Modeling,Shmueli2019Detecting,durmus2018exploring}, sequence ordering of arguments~\cite{Hidey2018Pesuasive}, and argument
structure features~\cite{LiDC20}.
The style feature is also proved to be significant in evaluating the persuasiveness of news editorial argumentation~\cite{Baff2020Analyzing}.
\citet{habernal2016what} conducted experiments in an entirely empirical manner, constructing a corpus for argument quality label classification and proposing several neural network models.

In addition to the features mentioned above, the role of pragmatic and discourse contexts has shown to be crucial by not yet fully explored.
\citet{Zeng2020What} examined how the contexts and the dynamic progress of argumentative conversations influence the comparative persuasiveness of an argumentation process.
\citet{durmus2019the} created a new dataset based on argument claims and impact votes from a debate platform \textit{kialo.com}, and experiments showed that incorporating contexts is useful to classify the argument impact.

Understanding discourse relations is one of the fundamental tasks of natural language understanding, and it is beneficial for various downstream tasks such as sentiment analysis~\cite{neja2017exploring,bhatia2015better}, machine translation~\cite{li2014assessing} and text generation~\cite{bosselut2018discourse}.
Discourse information is also considered indicative for various tasks of computational argumentation.
\citet{eckle2015on} analyzed the role of discourse markers for discriminating claims and premises in argumentative discourse and found that particular semantic group of discourse markers are highly predictive features. 
\citet{Hidey2018Pesuasive} concatenated sentence vectors with discourse relation embeddings as sentence features for persuasiveness prediction and showed that discourse embeddings helped improve performance.

%% file: sections/conclusion.tex
\section{Conclusion}

In this paper, we explicitly investigate how discourse structures influence the impact and the persuasiveness of an argument claim.
We present \textsc{DisCOC} to produce discourse-dependent contextualized representations.
Experiments and ablation studies show that our model improves its backbone RoBERTa around 1.67\%. Instead, HAN and other attention mechanisms bring side effects.
We discover distinct discourse relation path patterns and analyze representatives.
In the future, we plan to explore discourse structures in other NLU tasks.

%% file: sections/appendix.tex
\appendix


\section{Discourse Relation Path Patterns}
\label{appendix:path_pattern}
To explicitly explore important high-order discourse relation patterns, we model the process of yielding a concrete discourse relation path $p_{\text{disco}}(C^l) = (r^1, \cdots, r^l)$ as a generative process. For a given context path $(C^0, C^1, \cdots, C^{l-1})$ and the claim $C^l$, we define the \textbf{pattern set} as all possible patterns connected to $C^l$.
Mathematically, it is denoted as $\mathcal{P} = \sum_{i=1}^{l}{\bigtimes_{j=i}^{l}\mathcal{R}}$, where $\bigtimes$ is the Cartesian product.


We assume that every $r^i \in p_{\text{disco}}(C^l)$ is independent and identically distributed (i.i.d).
Under this assumption, the joint probability of a given path of discourse relations $(r^1, \cdots, r^l)$ is 
\begin{align}
    Pr(r^1, \cdots, r^l) = \Pi_{i=1}^l \bm{d}^i[r^i],
\end{align}
where $\bm{d}^i$ is the discourse relation distribution between $C^{i-1}$ and $C^i$, $\bm{d}^i[r^i]$ is the probability of a specific relation sense $r^i$.
Observing the consistently increased performance of BiLSTM on discourse relations in Figure~\ref{fig:lstm} when $l$ starts from 1 to 3 and no noticeable enhancement with longer contexts, we analyze path-generated distributions for up to three previous claims. 
We compute the joint probabilities $Pr(r^l), Pr(r^{l-1}, r^l), Pr(r^{l-2}, r^{l-1}, r^l)$ respectively and then concatenate these probabilities to get path pattern features $\bm{x} \in \mathbb{R}^{(|\mathcal{R}| + |\mathcal{R}|^2 + |\mathcal{R}|^3)}$ where each dimension of $\bm{x}$ corresponds to the probability of a pattern belonging to $\mathcal{P}$.
Next, the feature vector $\bm{x}$ is fed into a logistic regression (LR) model to train a one-vs-rest binary classifier for each of the three impact labels.

We report the largest top 10 coefficients of converged LR models in Table~\ref{table:disco_pattern}. 
Some relation path patterns are shown distinguishing for classifying individual impact labels.
Coefficients vary differently among different LRs except for {\textit{Restatement}-\textit{Restatement}}, which occurs in both \textit{Impactful} and \textit{Not Impactful}.
In general, \textit{Instantiation} is a typical relation that only occurs in the top patterns of \textit{Impactful}. Also, \textit{Restatement} is relatively frequent for \textit{Impactful} (5 of top 10), but it is the relation between the grandparent and the parent.
Providing additional resources (\textit{Restatement}-\textit{Result}) or objecting others' repetitions (\textit{Restatement}-\textit{Contrast}) can increase the persuasive power.
For the \textit{Medium Impact} class, its top 10 significant patterns are the longest on average.
That indicates some views are usually considered ordinary in complex structures.
\textit{Conjunction} is the dominant relation (8 of top 10) so that we are suggested to avoid to go along with others.
The case of \textit{Not Impactful} is a little clearer, in the sense that it has a unique relation \textit{Chosen Alternative} as one of the most significant patterns.
\textit{Restatement} also appears frequently, showing that neither generalization, nor specification, nor paraphrasing of others' views can help make claims stand out.
These interesting correlations between discourse relation path patterns and argument quality could be further analysis from the linguistic perspective in future works.


%% file: acl2021.bbl
\begin{thebibliography}{32}
\expandafter\ifx\csname natexlab\endcsname\relax\def\natexlab#1{#1}\fi

\bibitem[{Baff et~al.(2020)Baff, Wachsmuth, Khatib, and
  Stein}]{Baff2020Analyzing}
Roxanne~El Baff, Henning Wachsmuth, Khalid~Al Khatib, and Benno Stein. 2020.
\newblock Analyzing the persuasive effect of style in news editorial
  argumentation.
\newblock In \emph{ACL}, pages 3154--3160.

\bibitem[{Bhatia et~al.(2015)Bhatia, Ji, and Eisenstein}]{bhatia2015better}
Parminder Bhatia, Yangfeng Ji, and Jacob Eisenstein. 2015.
\newblock Better document-level sentiment analysis from {RST} discourse
  parsing.
\newblock In \emph{EMNLP}, pages 2212--2218.

\bibitem[{Bosselut et~al.(2018)Bosselut, Celikyilmaz, He, Gao, Huang, and
  Choi}]{bosselut2018discourse}
Antoine Bosselut, Asli Celikyilmaz, Xiaodong He, Jianfeng Gao, Po-Sen Huang,
  and Yejin Choi. 2018.
\newblock Discourse-aware neural rewards for coherent text generation.
\newblock In \emph{NAACL-HLT}, pages 173--184.

\bibitem[{Child et~al.(2019)Child, Gray, Radford, and Sutskever}]{ChildGRS2019}
Rewon Child, Scott Gray, Alec Radford, and Ilya Sutskever. 2019.
\newblock Generating long sequences with sparse transformers.
\newblock \emph{CoRR}, abs/1904.10509.

\bibitem[{Dai et~al.(2019)Dai, Yang, Yang, Carbonell, Le, and
  Salakhutdinov}]{DaiYYCLS19}
Zihang Dai, Zhilin Yang, Yiming Yang, Jaime~G. Carbonell, Quoc~Viet Le, and
  Ruslan Salakhutdinov. 2019.
\newblock Transformer-xl: Attentive language models beyond a fixed-length
  context.
\newblock In \emph{ACL}, pages 2978--2988.

\bibitem[{Devlin et~al.(2019)Devlin, Chang, Lee, and
  Toutanova}]{devlin2019bert}
Jacob Devlin, Ming{-}Wei Chang, Kenton Lee, and Kristina Toutanova. 2019.
\newblock {BERT:} pre-training of deep bidirectional transformers for language
  understanding.
\newblock In \emph{NAACL-HLT}, pages 4171--4186.

\bibitem[{Durmus and Cardie(2018)}]{durmus2018exploring}
Esin Durmus and Claire Cardie. 2018.
\newblock Exploring the role of prior beliefs for argument persuasion.
\newblock In \emph{NAACL-HLT}, pages 1035--1045.

\bibitem[{Durmus and Cardie(2019)}]{Durmus2019Modeling}
Esin Durmus and Claire Cardie. 2019.
\newblock Modeling the factors of user success in online debate.
\newblock In \emph{WWW}, pages 2701--2707.

\bibitem[{Durmus et~al.(2019)Durmus, Ladhak, and Cardie}]{durmus2019the}
Esin Durmus, Faisal Ladhak, and Claire Cardie. 2019.
\newblock The role of pragmatic and discourse context in determining argument
  impact.
\newblock In \emph{EMNLP-IJCNLP}, pages 5667--5677.

\bibitem[{Eckle-Kohler et~al.(2015)Eckle-Kohler, Kluge, and
  Gurevych}]{eckle2015on}
Judith Eckle-Kohler, Roland Kluge, and Iryna Gurevych. 2015.
\newblock On the role of discourse markers for discriminating claims and
  premises in argumentative discourse.
\newblock In \emph{EMNLP}, pages 2236--2242.

\bibitem[{Habernal and Gurevych(2016)}]{habernal2016what}
Ivan Habernal and Iryna Gurevych. 2016.
\newblock What makes a convincing argument? empirical analysis and detecting
  attributes of convincingness in web argumentation.
\newblock In \emph{EMNLP}, pages 1214--1223.

\bibitem[{Hidey and McKeown(2018)}]{Hidey2018Pesuasive}
Christopher Hidey and Kathleen~R. McKeown. 2018.
\newblock Persuasive influence detection: The role of argument sequencing.
\newblock In \emph{AAAI}, pages 5173--5180.

\bibitem[{Kingma and Ba(2015)}]{Kingma2014Adam}
Diederik~P. Kingma and Jimmy Ba. 2015.
\newblock Adam: {A} method for stochastic optimization.
\newblock In \emph{ICLR}.

\bibitem[{Lan et~al.(2017)Lan, Wang, Wu, Niu, and Wang}]{LanWWNW17}
Man Lan, Jianxiang Wang, Yuanbin Wu, Zheng{-}Yu Niu, and Haifeng Wang. 2017.
\newblock Multi-task attention-based neural networks for implicit discourse
  relationship representation and identification.
\newblock In \emph{EMNLP}, pages 1299--1308.

\bibitem[{Li et~al.(2020)Li, Durmus, and Cardie}]{LiDC20}
Jialu Li, Esin Durmus, and Claire Cardie. 2020.
\newblock Exploring the role of argument structure in online debate persuasion.
\newblock In \emph{EMNLP}, pages 8905--8912.

\bibitem[{Li et~al.(2014)Li, Carpuat, and Nenkova}]{li2014assessing}
Junyi~Jessy Li, Marine Carpuat, and Ani Nenkova. 2014.
\newblock Assessing the discourse factors that influence the quality of machine
  translation.
\newblock In \emph{ACL}, pages 283--288.

\bibitem[{Lin et~al.(2009)Lin, Kan, and Ng}]{LinKN09}
Ziheng Lin, Min{-}Yen Kan, and Hwee~Tou Ng. 2009.
\newblock Recognizing implicit discourse relations in the penn discourse
  treebank.
\newblock In \emph{ACL}, pages 343--351.

\bibitem[{Liu et~al.(2020)Liu, Ou, Song, and Jiang}]{liu2020bmgfroberta}
Xin Liu, Jiefu Ou, Yangqiu Song, and Xin Jiang. 2020.
\newblock On the importance of word and sentence representation learning in
  implicit discourse relation classification.
\newblock In \emph{IJCAI}, pages 3830--3836.

\bibitem[{Liu and Lapata(2019)}]{LiuL19}
Yang Liu and Mirella Lapata. 2019.
\newblock Text summarization with pretrained encoders.
\newblock In \emph{{EMNLP-IJCNLP}}, pages 3728--3738.

\bibitem[{Liu et~al.(2019)Liu, Ott, Goyal, Du, Joshi, Chen, Levy, Lewis,
  Zettlemoyer, and Stoyanov}]{liu2019roberta}
Yinhan Liu, Myle Ott, Naman Goyal, Jingfei Du, Mandar Joshi, Danqi Chen, Omer
  Levy, Mike Lewis, Luke Zettlemoyer, and Veselin Stoyanov. 2019.
\newblock Roberta: {A} robustly optimized {BERT} pretraining approach.
\newblock \emph{CoRR}, abs/1907.11692.

\bibitem[{Nejat et~al.(2017)Nejat, Carenini, and Ng}]{neja2017exploring}
Bita Nejat, Giuseppe Carenini, and Raymond Ng. 2017.
\newblock Exploring joint neural model for sentence level discourse parsing and
  sentiment analysis.
\newblock In \emph{SIGDIAL}, pages 289--298.

\bibitem[{Pennington et~al.(2014)Pennington, Socher, and
  Manning}]{Pennington2014Glove}
Jeffrey Pennington, Richard Socher, and Christopher~D. Manning. 2014.
\newblock Glove: Global vectors for word representation.
\newblock In \emph{EMNLP}, pages 1532--1543.

\bibitem[{Prasad et~al.(2008)Prasad, Dinesh, Lee, Miltsakaki, Robaldo, Joshi,
  and Webber}]{Prasd2008The}
Rashmi Prasad, Nikhil Dinesh, Alan Lee, Eleni Miltsakaki, Livio Robaldo,
  Aravind~K. Joshi, and Bonnie~L. Webber. 2008.
\newblock The penn discourse treebank 2.0.
\newblock In \emph{LREC}.

\bibitem[{Shmueli{-}Scheuer et~al.(2019)Shmueli{-}Scheuer, Herzig, Konopnicki,
  and Sandbank}]{Shmueli2019Detecting}
Michal Shmueli{-}Scheuer, Jonathan Herzig, David Konopnicki, and Tommy
  Sandbank. 2019.
\newblock Detecting persuasive arguments based on author-reader personality
  traits and their interaction.
\newblock In \emph{ACM-UMAP}, pages 211--215.

\bibitem[{Tan et~al.(2016)Tan, Niculae, Danescu{-}Niculescu{-}Mizil, and
  Lee}]{Tan2016Winning}
Chenhao Tan, Vlad Niculae, Cristian Danescu{-}Niculescu{-}Mizil, and Lillian
  Lee. 2016.
\newblock Winning arguments: Interaction dynamics and persuasion strategies in
  good-faith online discussions.
\newblock In \emph{WWW}, pages 613--624.

\bibitem[{Varia et~al.(2019)Varia, Hidey, and Chakrabarty}]{VariaHC19}
Siddharth Varia, Christopher Hidey, and Tuhin Chakrabarty. 2019.
\newblock Discourse relation prediction: Revisiting word pairs with
  convolutional networks.
\newblock In \emph{SIGdial}, pages 442--452.

\bibitem[{Vaswani et~al.(2017)Vaswani, Shazeer, Parmar, Uszkoreit, Jones,
  Gomez, Kaiser, and Polosukhin}]{VaswaniSPUJGKP17}
Ashish Vaswani, Noam Shazeer, Niki Parmar, Jakob Uszkoreit, Llion Jones,
  Aidan~N. Gomez, Lukasz Kaiser, and Illia Polosukhin. 2017.
\newblock Attention is all you need.
\newblock In \emph{NeurIPS}, pages 5998--6008.

\bibitem[{Wei et~al.(2016)Wei, Liu, and Li}]{wei2016post}
Zhongyu Wei, Yang Liu, and Yi~Li. 2016.
\newblock Is this post persuasive? ranking argumentative comments in online
  forum.
\newblock In \emph{ACL}, pages 195--200.

\bibitem[{Xue et~al.(2015)Xue, Ng, Pradhan, Prasad, Bryant, and
  Rutherford}]{XueNPPBR15}
Nianwen Xue, Hwee~Tou Ng, Sameer Pradhan, Rashmi Prasad, Christopher Bryant,
  and Attapol Rutherford. 2015.
\newblock The conll-2015 shared task on shallow discourse parsing.
\newblock In \emph{CoNLL}, pages 1--16.

\bibitem[{Yang et~al.(2019)Yang, Dai, Yang, Carbonell, Salakhutdinov, and
  Le}]{YangDYCSL19}
Zhilin Yang, Zihang Dai, Yiming Yang, Jaime~G. Carbonell, Ruslan Salakhutdinov,
  and Quoc~V. Le. 2019.
\newblock Xlnet: Generalized autoregressive pretraining for language
  understanding.
\newblock In \emph{NeurIPS}, pages 5754--5764.

\bibitem[{Yang et~al.(2016)Yang, Yang, Dyer, He, Smola, and Hovy}]{YangYDHSH16}
Zichao Yang, Diyi Yang, Chris Dyer, Xiaodong He, Alexander~J. Smola, and
  Eduard~H. Hovy. 2016.
\newblock Hierarchical attention networks for document classification.
\newblock In \emph{NAACL}, pages 1480--1489.

\bibitem[{Zeng et~al.(2020)Zeng, Li, He, Gao, Lyu, and King}]{Zeng2020What}
Jichuan Zeng, Jing Li, Yulan He, Cuiyun Gao, Michael~R. Lyu, and Irwin King.
  2020.
\newblock What changed your mind: The roles of dynamic topics and discourse in
  argumentation process.
\newblock In \emph{WWW}, pages 1502--1513.

\end{thebibliography}
